\documentclass[conference]{IEEEtran}
\usepackage{cite}
\usepackage{amsmath,amssymb,amsfonts}
\usepackage{algorithmic}
\usepackage{graphicx}
\usepackage{textcomp}
\usepackage{xcolor}
\usepackage{caption}
\usepackage{graphicx}
\usepackage{float} 
\usepackage{comment}
\usepackage{subcaption}
\usepackage{cleveref}
\usepackage[linesnumbered, ruled, noend]{algorithm2e}
\usepackage{amsmath}
\def\BibTeX{{\rm B\kern-.05em{\sc i\kern-.025em b}\kern-.08em
    T\kern-.1667em\lower.7ex\hbox{E}\kern-.125emX}}
\begin{document}

\title{AoI-Sensitive Data Forwarding with Distributed Beamforming in UAV-Assisted IoT
}

\author{
    \IEEEauthorblockN{
        Zifan Lang\IEEEauthorrefmark{2},
        Guixia Liu\IEEEauthorrefmark{2}\IEEEauthorrefmark{1},
        Geng Sun\IEEEauthorrefmark{2}\IEEEauthorrefmark{3}\IEEEauthorrefmark{1},
        Jiahui Li\IEEEauthorrefmark{2},
        Zemin Sun\IEEEauthorrefmark{2},
        Jiacheng Wang\IEEEauthorrefmark{3},
        Victor C.M. Leung\IEEEauthorrefmark{4}\IEEEauthorrefmark{6}
    }
    \IEEEauthorblockA{\IEEEauthorrefmark{2}College of Computer Science and Technology, Jilin University, Changchun 130012, China}
    \IEEEauthorblockA{\IEEEauthorrefmark{3}College of Computing and Data Science, Nanyang Technological University, Singapore 639798, Singapore}
    \IEEEauthorblockA{\IEEEauthorrefmark{4}Artificial Intelligence Research Institute, Shenzhen MSU-BIT University, Shenzhen 518115, China}
    \IEEEauthorblockA{\IEEEauthorrefmark{6}Department of Electrical and Computer Engineering, The University of British Columbia, Vancouver V6T 1Z4, Canada}
    \IEEEauthorblockA{\IEEEauthorrefmark{1}Corresponding author: Guixia Liu and Geng Sun}
}
\maketitle

\begin{abstract}
This paper proposes a UAV-assisted forwarding system based on distributed beamforming to enhance age of information (AoI) in Internet of Things (IoT). Specifically, UAVs collect and relay data between sensor nodes (SNs) and the remote base station (BS). However, flight delays increase the AoI and degrade the network performance. To mitigate this, we adopt distributed beamforming to extend the communication range, reduce the flight frequency and ensure the continuous data relay and efficient energy utilization.
Then, we formulate an optimization problem to minimize AoI and UAV energy consumption, by jointly optimizing the UAV trajectories and communication schedules.
The problem is non-convex and with high dynamic, and thus we propose a deep reinforcement learning (DRL)-based algorithm 
to solve the problem, thereby enhancing the stability and accelerate convergence speed. Simulation results show that the proposed algorithm effectively addresses the problem and outperforms other benchmark algorithms.
\end{abstract}


\section{Introduction}

\par Internet of Things (IoT) have been proven to be highly effective in remote areas, particularly for the tasks such as monitoring, sensing, and detection, where data is rapidly transmitted to fusion centers. To automate data integration from these deployed sensor nodes (SNs), UAVs are deployed to operate between IoT and the BS to facilitate data collection, dissemination, and relay to enhance network efficiency~\cite{Chang2024}. Consequently, UAVs play a crucial role in ensuring seamless data flow from remote IoT to central processing units, which is especially beneficial in challenging environments.

\par Despite the advantages of UAV-assisted IoT, a significant challenge arises when timely data delivery is required. Specifically, the flight of UAVs introduces delays in data reception, which leads to a high age of information (AoI) and reduces real-time system responsiveness~\cite{Long2022}. This issue becomes especially critical in the applications such as forest fire monitoring and border surveillance, where the delayed information results in severe consequences~\cite{Gao2023}. Therefore, minimizing the UAV flight time while maintaining efficient data transmission to reduce AoI remains an urgent task. Addressing this challenge demands innovative approaches that ensure both timely data delivery and efficient resource utilization.

\par Distributed beamforming offers a promising solution to these challenges by significantly enhancing the transmission gain and extending the communication range~\cite{Li2024}. By employing distributed beamforming, UAVs can maintain connectivity with the BS without frequent flights, thereby mitigating flight-induced delays and reducing AoI. This capability is particularly valuable in the scenarios requiring continuous data relay, as it allows for more efficient use of UAV energy resources.

\par Designing an effective UAV-assisted relay system involves several complexities. First, distributed beamforming introduces multiple transmission phases, which increases the difficulty of system modeling. Additionally, optimizing UAV trajectories is essential for minimizing both AoI and energy consumption across sensors. Furthermore, the uncertainty and variability introduced by dynamic updates of sensor data demand robust optimization strategies that can adapt in real time. However, addressing these complexities requires innovative modeling and optimization techniques that can effectively handle the dynamic and multi-faceted nature of the problem.

\par Unlike previous studies, this paper introduces a UAV-assisted AoI-sensitive data forwarding system and proposes a modified deep reinforcement learning (DRL) algorithm to minimize the AoI of SNs and energy consumption of UAVs. The primary contributions of this work are outlined as follows:

\begin{itemize}
    \item \textit{UAV-assisted AoI-sensitive Data Forwarding System:}
    We consider a UAV-assisted AoI-sensitive data forwarding system with distributed beamforming. Specifically, the direct link from the SNs to BS is blocked due to the long-range transmission distance, and distributed beamforming is employed to mitigate flight-induced delays of UAVs. 
    \item \textit{Formulation of Complex Optimization Problem:}
    In the considered system, we construct a joint optimization problem that cooperatively plans the UAV trajectory and schedules communication to minimize the AoI of SNs and the energy consumption of UAVs. The real-time requirements and dynamic nature of this problem make it challenging to be solved efficiently.
    \item \textit{Enhanced DRL-based Approach:}
    We propose a SAC with temporal sequence, layer normalization gated recurrent unit, and attention (SAC-TLA), which is a DRL-based algorithm to solve our optimization problem. Specifically, SAC-TLA introduces temporal sequence, layer normalization gated recurrent unit (LNGRU), and attention mechanism to enhance the stability and accelerate convergence speed. 
    \item \textit{Performance Evaluation:}
    Simulations results present that the proposed SAC-TLA approach achieves superior performance compared to other benchmarks.
\end{itemize}

\par The rest structure of this paper is organized as follows. Section II presents the models and formulates the optimization problem. Section III introduces the proposed SAC-TLA algorithm. Simulation results are shown in section IV. Finally, section V provides the conclusions.
\begin{figure}[t]
    \centerline{\includegraphics[width=3.5in]{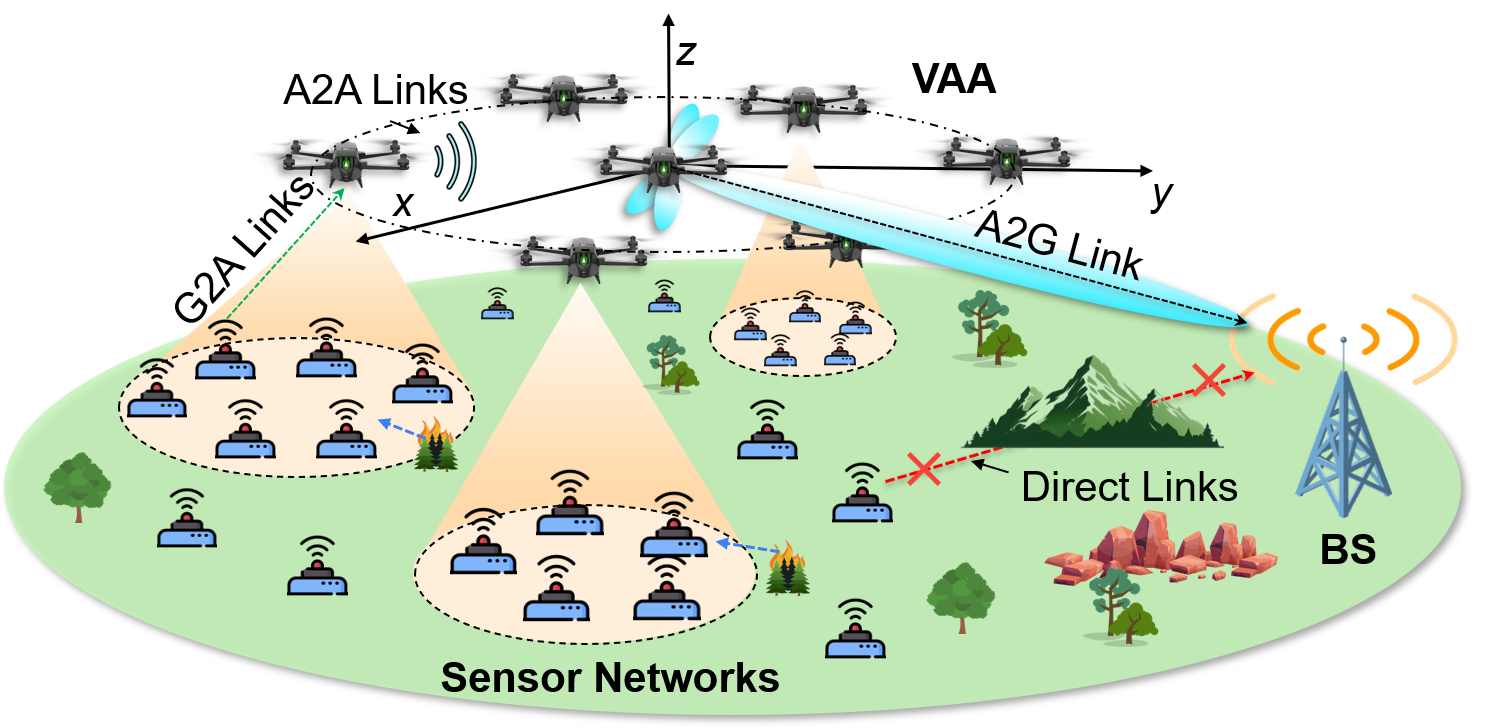}}
    \caption{A UAV-assisted AoI-sensitive data forwarding system in IoT based on distributed beamforming.}
    \label{fig:system model}
 \end{figure}   

\section{System Models}
\par In this section, we first present the system overview. Subsequently, we introduce the considered models, which includes the network, AoI, and UAV energy cost models, to characteristize the optimization objectives and decision variables. 
\vspace{-0.6em} 
\subsection{System Overview}
\vspace{-0.6em} 
As shown in Fig.~\ref{fig:system model}, we consider a UAV-assisted AoI-sensitive data forwarding system in IoT. In this system, the SNs denoted as $\mathcal{N}\triangleq \{i\mid 1,2,\dots,N_{SN}\}$ are randomly distributed across the monitor area. We consider that each SN generates data in each time slots and the volume of data for each SN is denoted by $\mathcal{D}\triangleq \{D_{1},D_{2},\dots,D_{N_{SN}}\}$. Due to the complex terrestrial network environments and long distance between these SNs and BS, the direct transmissions between them may be infeasible. Under the circumstances, a fleet of UAVs denoted as $\mathcal{U}\triangleq \{j\mid 1,2,\dots,N_{UAV}\}$ are deployed to collect and forward the data from the sensor network to the remote BS. Note that each UAV is equipped with a single omnidirectional antenna. In order to reduce the information delay caused by the UAV flying back and forth between the SNs and the BS, we consider introducing distributed beamforming method and constructing the UAVs as a VAA to enhance the transmission ranges without moving UAVs, thereby reducing the AoI of these SNs. 

\par
\begin{figure}[t]
    \centerline{\includegraphics[width=3in]{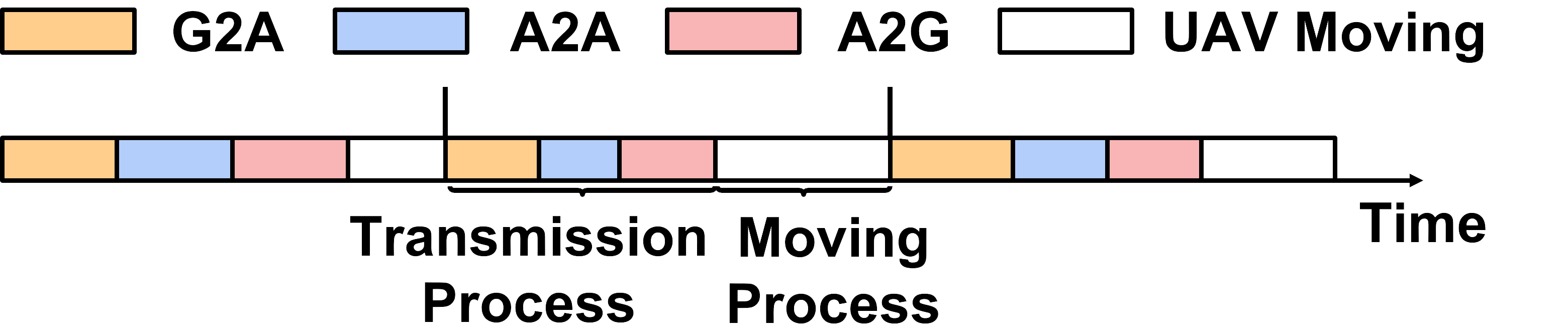}}
    \caption{Time allocation for transmission and moving processes.}
    \label{fig:time slot}
 \end{figure}

 Based on this, we consider a time-slotted muti-access protocol, which is shown in Fig.~\ref{fig:time slot}. Specifically, each frame is divided into multiple time slots with a unit length. As such, the set of all time slots is denoted by $\mathcal{T}\triangleq \{t\mid 1,2,\dots,T\}$. Following this, each time slot is divided into four phases, $\delta_{G2A}(t)$ for G2A transmission, $\delta_{A2A}(t)$ for A2A transmission, $\delta_{A2G}(t)$ for A2G transmission, and $\delta_{move}(t)$ for the UAV moving process. During the G2A transmission phase, each UAV collects sensing data from SNs within its coverage area. In the A2A transmission phase, the UAVs broadcast the collected data to other UAVs. In the A2G transmission phase, the UAVs form a VAA to communicate with the BS using distributed beamforming. Finally, in the UAV moving phase, the UAVs fly to their next designated hovering positions.

\par For generality, we adopt a 3D Cartesian coordinate system, where the position of the $i$-th SN is $\boldsymbol{q}_{i}^{SN} = (x{i}, y_{i}, 0)$, and the position of the $j$-th UAV in time slot $t$ is $\boldsymbol{q}_{j}^{UAV}(t) = (x{j}(t), y_{j}(t), H^U)$. Subsequently, we will detail the network model, which contains three key models with respect to transmission, i.e., the G2A, A2A, and A2G transmission models. Following this, we will derive an AoI model and an UAV energy cost model. 
 \vspace{-0.6em} 
\subsection{Network Model} 
\vspace{-0.6em} 
\par In this section, we present a detailed introduction of the G2A, A2A, and A2G transmission models.

\par {\textit{1) G2A Transmission Model.}}
In this part, we apply the frequency division multiplexing access (FDMA) mechanism and calculate the transmission rate using probabilistic LoS as the communication link between the SNs and UAVs. Specifically, we consider that different SNs utilize different frequencies to prevent interference during data transmission. Besides, we assume that a SN can only communicate with one UAV at a time, while a UAV can communicate with multiple SNs simultaneously. To this end, we define a binary variable $\beta_{i,j}(t)$ to represent whether the $j$-th UAV communicates with the $i$-th SN in time slot $t$. Therefore, the communication scheduling constraints for SNs are as follows:
\begin{align}
    \sum\nolimits_{j = 1}^{N_{UAV}} \beta_{i,j}(t)\leq 1,\beta_{i,j}(t)\in \{0,1\},\forall i\in\mathcal{N}, \forall j\in\mathcal{U} . \label{beta}
\end{align}

\par Then, we consider a channel model between the $i$-th SN and the $j$-th UAV, which includes both line-of-sight (LoS) and non-line-of-sight (NLoS) links. The probability of LoS is $P_{LoS}(\theta_{i,j}(t))=1/[1+a \exp(-180 b(\theta_{i,j}(t))/\pi-a)]$,
where $a$ and $b$ are two constants related to the environment, and $\theta_{i,j}(t)=\sin^{-1}(h_{j}^{UAV}(t)/d_{i,j}^{H})$, where $h_{j}^{UAV}(t)$ and $d_{i,j}^{H}$ are the vertical and horizontal distance between the $j$-th UAV and the $i$-th SN, respectively. Correspondingly, the probability of NLoS is $P_{NLoS}=1-P_{LoS}(\theta)$. Furthermore, we support that the conditions of the uplink and downlink channels between SNs and UAVs are comparable. In this case, the average channel power gain between the $i$-th SN and the $j$-th UAV can be expressed as $g_{i,j}=[P_{LoS}L_{LoS}+P_{NLoS}L_{NLoS}]^{-1}$. Then, the uplink data transmission rate between the UAV and the SN is given by $R_{i,j}=B_{i,j}\log_{2}(1+P_{i}g_{i,j}/{\sigma^{2}})$, where $P_{i}$ is the transmit power and $\sigma^{2}$ is the noise power.

\par By collecting data from accessible SNs, the capacity of the $j$-th UAV during the G2A transmission is given by 
\begin{align}
    S_{j}=\sum\nolimits_{i=1}^{N_{SN}} \beta_{i,j}D_{i}.\label{su}
\end{align}
Then, the transmission time of $j$-th UAV at time slot $t$ is denoted as $\delta_{j}(t)=\sum_{i=1}^{N_{SN}} (\beta_{i,j}D_{i})/R_{i,j}$. Thus, the phase of G2A transmission is $\delta_{G2A}(t)=\max \left( \delta_j(t) \mid 1 \leq j \leq N_{\text{UAV}} \right)$.

\par{\textit{2) A2A Transmission Model.}
Upon collecting complete data from a SN, the UAV receiver will broadcast the data to all UAVs. Here, we consider that the UAV adopts FDMA to broadcast data to other UAVs. Given the high operating altitude of UAVs, a LoS channel model should be used for A2A transmission. We use $d_{j,j'}$ to indicate the distance between the $j'$-th UAV and the $j$-th UAV, and the channel power gain between $j$-th UAV and $j'$-th UAV can be expressed by $h_{j,j'}=\rho _{0}/{d_{j,j'}^{\alpha}}$, where parameter $\rho _{0}$ is the channel power gain associated with a reference distance and parameter $\alpha$ is the path loss exponent. Then, the transmission rate during the aerial broadcast phase is given by
\begin{align}
    R_{j,j'}^{A2A} = B_{j'} \log_{2} \left( 1 + P_{j'}h_{j,j'} /{ \sigma^{2}} \right),
\end{align}
where parameter $B_{j'}$ is the channel bandwidth allocated to $j'$th UAV and $P_{j'}$ is the transmit power of $j'$-th UAV. Due to the characteristics of broadcasting, the actual broadcast rate of the $j$-th UAV should be the minimum rate $R_{j}^{A2A}$ achievable by all UAVs to ensure simultaneous data reception. 

\par Since the distances between UAVs to each other are much smaller than the distances between UAVs and BS, we assume that all data collected from SNs can be reliably broadcast to the UAV swarm within each time slot. 
In this case, the phase of A2A transmission is given by $\delta_{A2A}(t)=\max \left( S_{j}/{R_{j}^{A2A}} \mid 1 \leq j \leq N_{\text{UAV}} \right)$, 
where $S_{j}$ is data successfully collected by the $j$-th UAV which is denoted in Eq. (\ref{su}).

\par\textit{3) A2G Transmission Model Based on UAV-Enabled VAA.}
After the data broadcast to all UAVs, we use distributed beamforming for data transmission between UAVs and the BS. We consider the channel condition between the VAA and the BS is LoS. Then, the signal-to-noise ratio (SNR) of the BS can be expressed as $\gamma_{SNR}(t)=(\sum_{\forall j\in \mathcal{U}} \sqrt{P_{j}(t)g_{0}d_{j,BS}^{-\alpha}})^2/{\sigma^2}$~\cite{Li2024a}, where $P_{j}(t)$ represents the transmit power at time slot $t$, $g_{0}$ represents the channel power gain, $d_{j,BS}$ denotes the propagation distance between $j$-th UAV and BS, $\alpha$ is the path loss exponent. Based on this, the achievable rate from the VAA to the BS is given by
\begin{align}
    R_{BS}(t)=B\log_{2}(1+\gamma_{SNR}(t)),
\end{align}
where $B$ is the carrier bandwidth. Then, the data successfully forwarded from the VAA to the BS in time slot $t$ can be denoted by $S_{A2G}=\delta_{A2G}(t)R_{BS}(t)$, 
where $\delta_{A2G}(t) \leq 1-\delta_{G2A}(t)-\delta_{A2A}(t)-\delta_{move}(t)$. Note that distributed beamforming is to calculate the maximum capacity of the VAA to the BS without optimization. 
\vspace{-0.5em} 
\subsection{AoI Model}
\vspace{-0.5em} 
In this work, data validity is closely associated with the timeliness of its collection and transmission. Thus, we adopt the concept of AoI $A_{i}(t)$ to reflect the freshness of the data from $i$-th SN at time slot $t$. 

\par If the $i$-th SN fails to communicate with any UAV in time slot $t$, its data incurs an additional delay, and the AoI updates as $A_{i}(t+1) = A_{i}(t) + 1$. If the $j$-th UAV broadcasts to all UAVs and forwards the data from the $i$-th SN to the BS, the AoI of the $i$-th SN will decrease in the following time slot.

\par Due to limited channel capacity, the VAA may not always successfully forward data to the BS during the A2G phase. In this case, the fraction of data successfully forwarded is $Q(t)\triangleq \min\{S_{G2A},S_{A2G}\}/{S_{c}}$, where $S_{c}=\sum_{i=1}^{N_{SN}} D_{i}$ is the size of sensing data of all SNs before data collection
and $S_{G2A}=\sum_{j=1}^{N_{UAV}} S_{j}$ is the size of received data during G2A transmission, where $S_{j}$ is shown in Eq. \eqref{su}. Thus, the AoI dynamics of SNs can be summarized as follows: 
\begin{align}
    A_{i}(t+1)=
    \begin{cases}
        (1-Q(t))(A_{i}(t)+1), \text{if}\enspace\beta_{i,j}(t)=1,\\
        A_{i}(t)+1,\quad\text{otherwise.}\label{con:inventoryflow} 
    \end{cases}
\end{align} 
Then, the time-averaged AoI is defined as 
\begin{align}
    A = \frac{1}{T} \frac{1}{N} \sum\nolimits_{t=1}^{T} \sum\nolimits_{i=1}^{N_{SN}} A_i(t).\label{aoi}
\end{align}
\vspace{-0.8em} 
\subsection{UAV Energy Cost Model}
\vspace{-0.5em} 
In this section, we detail the following energy consumption model for UAVs. Specifically, the main energy consumption of the UAV is propulsion energy consumption, since the communication-related energy is two orders of magnitude smaller than the propulsion energy of UAV. Thus, for an altitude-fixed rotary-wing UAV, we denote the propulsion power as $P_{UAV}(v)$, which follows the model in~\cite{Zeng2019}.
\par Thus, the flight energy consumption and the hovering energy consumption of $j$-th UAV are given by $E_{j}^{move}=\sum\nolimits_{t=1}^{T} P_{UAV}(v)\delta_{move}(t)$ and $E^{hov}_{j}=\sum\nolimits_{t=1}^{T} P_{hov}(\delta_{G2A}(t)+\delta_{A2A}(t)+\delta_{A2G}(t))$, respectively. Then, the energy consumption of UAVs can be denoted as $E=\sum\nolimits_{j=1}^{N_{UAV}}E_{j}^{move}+E^{hov}_{j}$. 
Note that we omit the extra energy consumption caused by the acceleration and deceleration of the UAV during horizontal flight, since it constitutes only a minor part of the total operation time in the maneuver duration of the UAV. 
\vspace{-0.5em} 
\subsection{Problem Formulation}
\vspace{-0.5em} 
\par We formulate the optimization problem for the UAV-assisted AoI-sensitive data forwarding system based on the models mentioned above.
Specifically, our objective is to minimize the time-average AoI of SNs while minimizing the energy consumption of the UAVs. Thus, we jointly optimize the trajectory of the UAVs and the communication schedule for SNs in each time slot. Accordingly, the optimization problem can be formulated as:
\begin{subequations}\label{opti}
\begin{align}
\mathop{\min}\limits_{\mathbf{\Phi},\mathbf{Q}} 
 & \ (A+E), \\
\text{ s.t. }
    &\beta_{i,j}(t)\in \{0,1\} \forall i\in \mathcal{N},\forall j\in \mathcal{U},\forall t\in \mathcal{T},\label{opti:sub1}\\
    &(x_{j}^{UAV},y_{j}^{UAV},z_{j}^{UAV})\in \mathbb{R}_{j}^{3\times 1} , \quad \forall j\in \mathcal{U}, \label{opti:sub2}\\
    &d_{j1,j2}\geqslant d_{min},\label{opti:sub3}\\
    &\|\boldsymbol{q}_{j}^{UAV}(t+1)-\boldsymbol{q}_{j}^{UAV}(t)\|^{2}\leq (v_{max}\delta_{move}(t))^{2},
\end{align}
\end{subequations}
where $\{\mathbf{\Phi} = {\beta_{i,j}(t)},\forall i\in \mathcal{N},\forall j\in \mathcal{U},\forall t\in \mathcal{T}\}$ represents the binary variable of communication schedule for the SNs and $\{\mathbf{Q}=\boldsymbol{q}^{UAV}(t),\forall t\in \mathcal{T}\} $ denotes the position of UAVs. This problem is non-convex referring to~\cite{Long2022} and it is generally quite difficult to solve in real-time. 

\section{The Proposed SAC-TLA}
\par In this section, we propose a DRL-based method to solve our optimization problem. To this end, we first show the motivations for using DRL and reformulate our problem as a markov decision process (MDP). Then, we introduce the proposed SAC-TLA algorithm with several improvements.
\vspace{-0.3em} 
\subsection{Motivations and MDP}
\vspace{-0.3em} 
In this part, we first introduce the motivations for using DRL, and then present the formulation and simplification of the MDP. 
\par\textit{1) Motivations for Using DRL.}
Our problem is dynamic and uncertain, with factors like UAV mobility and changes over multiple transmission phases. Thus, the commonly used static optimization methods such as convex or non-convex optimization are not suitable for this problem~\cite{Li2024}. In this case, DRL is able to swiftly adapt the dynamic and uncertain environment and offer robust solutions. Thus, we seek to adopt the DRL method to solve our optimization problem. 

\par\textit{2) MDP Formulation.}
We reformulate the optimization problem shown in Eq. \eqref{opti} as an MDP. Mathematically, an MDP is a tuple \( (S, A, P, R, \gamma) \) which are state space, action space, state transition probability, reward function, and discount factor, respectively. Among them, state, action, and reward are the most important components, which are detailed as follows.

\begin{itemize}
    \item \textbf{State Space:} The state space is designed to capture essential spatial and operational dynamics influencing system performance. Specifically, the positions of both UAVs and SNs, along with the AoI of each SN, are incorporated to reflect spatial dynamics. Thus, the state at time slot $t$ is formally expressed as $s_t = \{ \boldsymbol{q}^{UAV}_t, \boldsymbol{q}^{SN}_t,\boldsymbol{A}^{SN}_t\}$.
    \item \textbf{Action Space:} In our system, the UAV can adjust its position to optimize its communication with SNs. Additionally, the binary variable $\beta_{i,j}(t)$ reflects the communication scheduling decision, which determines whether the $j$th UAV communicates with the $i$th SN in time slot $t$. As such, the action of agent at time slot $t$ is represented as $a_j(t) = \{a_j^{x}(t), a_j^{y}(t)\}$, where $a_j^{x}(t)$ and $a_j^{y}(t)$ denote the movement of $j$-th UAV in the $x$ and $y$ directions at time slot $t$, respectively.
    \item \textbf{Reward Function:} A well-designed reward function contributes to problem-solving and is critical. As such, the reward function involves our optimization objective and the corresponding constraints shown in Eq. \eqref{reward}. Specifically, our reward of $j$-th agent is defined as: 
\begin{align}
    r_j(t)= -\rho_1A-\rho_2E+\rho_3c_j-p_j, \label{reward}
\end{align}
where $\rho_1$, $\rho_2$, $\rho_3$ are three normalization parameters to adjust to bring these terms to the same order of magnitude, $c_j$ is the amount of SNs covered by the $j$-th UAV to guide agents to avoid the impact of extreme negative rewards, and $p_j$ is the out-of-bounds penalty.

\end{itemize}  
\par\textit{3) MDP Simplification.}
To efficiently handle the complexity arising from the mixed discrete and continuous action space, we simplified MDP. Specifically, the discrete nature of communication actions introduces considerable randomness into the decision making process. This randomness complicates the decision making and learning process, and leads to inefficiencies. As a consequence, the selected SNs may fall outside the communication range of UAVs, while SNs within the radio range remain unselected, which leads to wasted communication resources.

\par To resolve these issues, we introduce dynamic proximity-based action mapping (DPAM), a strategy that simplifies decision-making by directly linking UAV trajectories to communication actions~\cite{Sun2024a}. In particular, DPAM makes communication decisions deterministically by checking whether SNs are within the communication radius of the UAV, thereby eliminating the unpredictability associated with binary actions. This reformulation ensures reliable data collection, enhances energy efficiency, and streamlines the action space. 
\vspace{-0.3em} 
\subsection{Standard SAC Algorithm}
\vspace{-0.3em} 
In this paper, we adopt the soft actor-critic (SAC) algorithm as our optimization framework~\cite{Luong2019}. SAC is a model-free, off-policy reinforcement learning method, which is suitable for high-dimensional and continuous action spaces, and the objective of this method is to maximize the discounted rewards and policy entropy, encourages exploration, and avoids premature convergence to a deterministic policy, i.e.,
\begin{align}
    \max_{\theta} J(\theta) = \mathbb{E}_{\pi_\theta} \left[ \sum\nolimits_{t=1}^{T} \gamma^t (r_t + \alpha \mathcal{H}(\pi(\cdot|s_t))) \right].
\end{align}

\par Moreover, SAC consists of an actor network \( \pi_\theta \), which produces a stochastic policy, and two critic networks \( Q_\phi \), which is used to mitigate overestimation by minimizing the Bellman error, i.e.,
\begin{align}
\begin{split}
L_Q(\phi) = \mathbb{E} \Big[ \Big( Q_\phi(s_t, a_t) &- \Big( r_t + \gamma \Big( Q_{\bar{\phi}}(s_{t+1}, a_{t+1}) \\
&- \alpha \log \pi_\theta(a_{t+1} | s_{t+1}) \Big)\Big)\Big)^2 \Big],
\end{split}
\end{align}
\par However, in our MDP, the reward function is highly variable, which makes it difficult for the standard SAC to quickly adjust policy and results in slow convergence. Therefore, we propose several enhancements to SAC tailored to our scenario.
\vspace{-0.6em} 
\subsection{SAC-TLA Algorithm} 
In this work, we introduce an enhanced SAC-TLA. Specifically, SAC-TLA integrates three essential enhancements which are temporal sequence input processing, LNGRU, and attention mechanism, and they are detailed as follows.

\par\textit{1) Temporal Sequence Input Processing.} To improve the learning capability of conventional SAC algorithm in a time-varying environment, we introduce the temporal sequence-based state input. Unlike conventional SAC, which processes each state individually, temporal sequence input allows the algorithm to leverage temporal information from multiple time steps. Specifically, the environment generates a sequence of states $\{s_t, s_{t-1}, ..., s_{t-n}\}$ as the input to actor and critic networks, which enables the agents to better capture dependencies and patterns in the environment to improve policy performance ultimately.
\begin{figure}[t]
    \centerline{\includegraphics[width=3.4in]{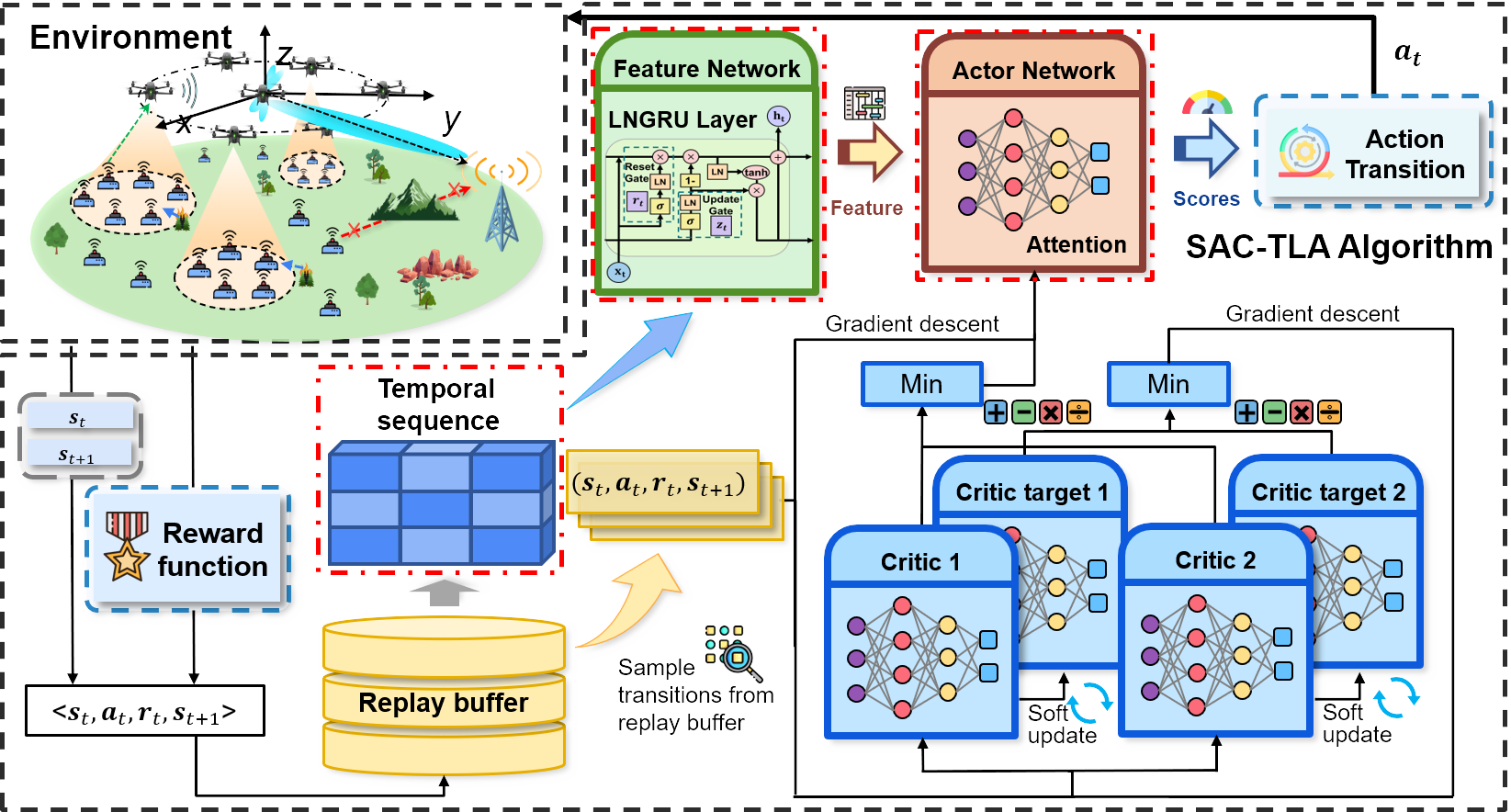}}
    \caption{The framework of the SAC-TLA algorithm.}
    \label{fig:SAC-TLA}
\end{figure} 
\par\textit{2) LNGRU.}
To capture long-term dependencies in sequential data, we integrate LNGRU cells into actor network~\cite{LeiBa2016}. Unlike standard GRU, LNGRU applies layer normalization after the reset gate \( r_t \), update gate \( z_t \), and candidate hidden state \( \tilde{h}_t \) to stabilize gradient flow. Specifically, the candidate hidden state \( \tilde{h}_t \) is expressed as:
\begin{align}
    \tilde{h}_t = \tanh(LN(W_h x_t) + r_t \circ LN(U_h h_{t-1})),
\end{align}
Following this, the normalization mitigates gradient vanishing and explosion, and ensures efficient backpropagation over time. Thus, LNGRU enhances the ability of the algorithm to learn temporal dependencies and improves decision making in dynamic environments.

\par\textit{3) Attention Mechanism.}
To enhance the focus of algorithm on critical elements within input sequences, we integrate a global soft attention mechanism into the LNGRU cell~\cite{Brauwers2023}. Specifically, the attention assigns varying weights to the LNGRU hidden states, to emphasize key features at each time step. For each hidden state \( h_t \), the attention score \( \alpha_t \) is expressed as:
\begin{align}
    \alpha_t = \exp(h_t^\top W_a h_{c})/{\sum\nolimits_{t'} \exp(h_{t'}^\top W_a h_{c})},
\end{align}
where \( W_a \) is the attention weight matrix, and \( h_{c} \) summarizes global sequence information. Then, the attention-weighted hidden state \( \hat{h}_t \) is fed into the actor and critic networks to direct the focus of agent toward essential input features. The synergy between attention and LNGRU improves decision-making and performance in dynamic environments.

\par\textit{4) SAC-TLA Framework.} As shown in Fig.~\ref{fig:SAC-TLA}, SAC-TLA extends the core structure of SAC by approximating the policy and the value function through actor and critic networks, respectively. Specifically, the actor network combines LNGRU and attention mechanism to generate action \(a_t\) based on the state sequence. Moreover, the actor network combines LNGRU and the attention mechanism to generate an action \(a_t\) based on the state sequence. Then, the critic network estimates the Q-value of the state-action pair. Each interaction transfer \( (s_t, a_t, r_t, s_{t+1})\) is stored in a replay buffer from which a small batch of samples is drawn to update the network during training. In addition, the main steps of the proposed SAC-TLA algorithm are shown in Algorithm~\ref{alg:lasac}.
\setlength{\headsep}{25pt}  
\begin{algorithm}[t]
\caption{SAC-TLA Algorithm}\label{alg:lasac}
     Initialize actor network parameters $\theta$ and critic network parameters $\phi$ \;
     Initialize target critic network $\phi_{\text{target}} \gets \phi$\;
     Initialize attention weights $W_{\text{att}}$\;
     Initialize entropy coefficient $\alpha$ and log $\alpha$\;

    \For{each episode}
        { Reset environment and initialize state $s_t$\;
        \For{each time slot $t$}
            { Obtain state sequence $\{s_{t-n}, ..., s_{t-1}, s_t\}$\;
             Compute hidden state $h_t$ using LNGRU Cell from $\{s_{t-n}, ..., s_t\}$\;
             Apply attention to compute features $\hat{h}_t = \text{Attention}(h_t, W_{\text{att}})$\;
             Select action $a_t$ with exploration noise\;
             Calculate AoI $A$ and energy consumption $E$\; 
             Update reward $r_j$ according to Eq. (\ref{reward})\;
             Store transition $\{s_t, a_t, r_t, s_{t+1}\}$ into replay buffer\;
        }
        \If{replay buffer size $\geq$ batch size}
            {\For{each update step}
                { Sample a batch of transitions $\{s, a, r, s'\}$ \;
                 Compute target $Q$ values and update critic network\;
                 Update actor network\;
                 Soft update target network: 
                $\phi_{\text{target}} \gets \tau \phi + (1 - \tau) \phi_{\text{target}}$\;
            }
        }
    }
\end{algorithm}

\begin{figure*}[t]
    \centerline{\includegraphics[width=\linewidth]{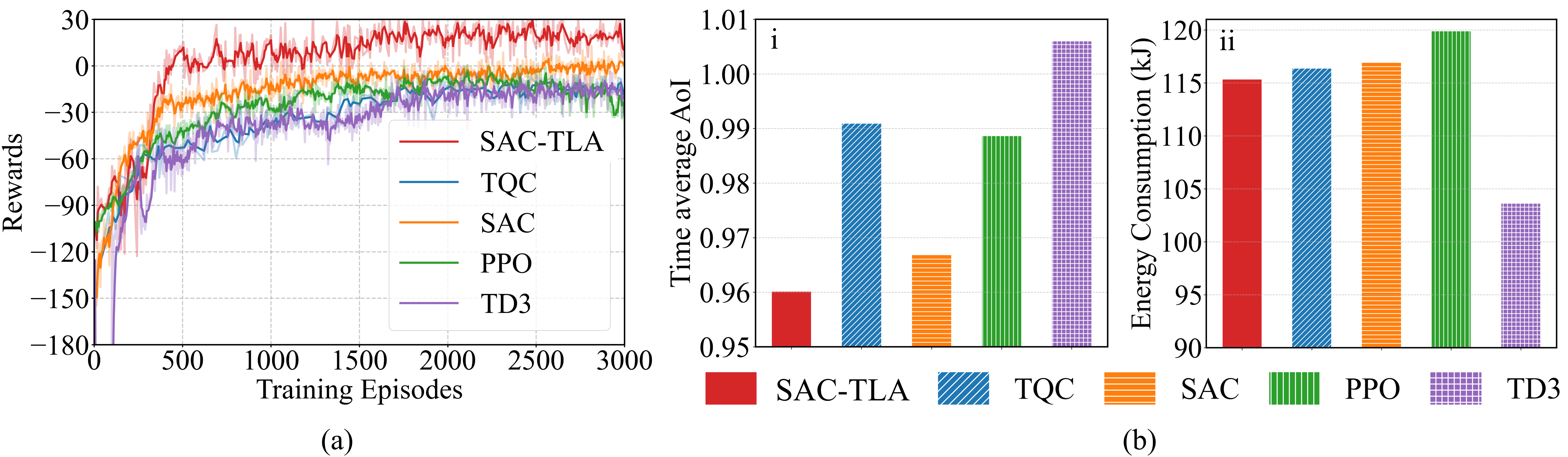}}
    \caption{Simulation results. (a) Cumulative rewards training curve. (b) The optimization objective values of SAC-TLA, TQC, SAC, PPO, and TD3.}
    \label{fig:Simulation results}
\end{figure*} 
\section{Simulation Results and Analysis}
\par In this section, we detail the simulation results and analyses. We first introduce the simulation setting and benchmarks, and then provide the corresponding results.

\par Our coordinate system places the sensing area left corner as the origin, with BS coordinates at (1000, 1000, 0). We model the sensing area as a square region measuring 400 m $\times$ 400 m, with 60 SNs randomly deployed across it. Following this, 4 UAVs are deployed randomly in a valid monitor area for each episode.
For comparison, we utilize twin delayed deep deterministic policy gradient (TD3)~\cite{Luong2019}, proximal policy optimization (PPO)~\cite{Luong2019}, truncated quantile critics (TQC)~\cite{Luong2019} and soft actor-critic (SAC)~\cite{Luong2019} as benchmark methods.  

\par Fig.~\ref{fig:Simulation results}(a) illustrates the cumulative rewards for each episode of SAC-TLA in comparison to other benchmark algorithms. As can be seen, the SAC-TLA algorithm outperforms other algorithms with higher rewards, faster convergence, and better stability. The reason may be attributed to following factors. First, our temporal sequence input enables the algorithm to capture evolving environment dynamics. This allows the UAVs to anticipate changes and make proactive decisions. Second, LNGRU stabilizes learning by effectively handling long-term dependencies and preventing gradient issues. This ensures efficient policy updates over extended time horizons. Finally, the global soft attention mechanism focuses on SNs with highly AoI, and optimizes UAV actions ro reduce unnecessary exploration. In this case, these components enable SAC-TLA to learn more efficiently, which leads to faster convergence by streamlining decision-making and optimizing both data collection and energy consumption. 

\par Fig.~\ref{fig:Simulation results}(b) provides the comparative results of time average AoI of SNs and UAVs energy consumption among five methods. We can observe that the SAC-TLA algorithm achieves the lowest AoI while maintaining competitive energy consumption. This indicates the effectiveness of SAC-TLA in balancing timely data collection with efficient resource utilization. In contrast, TD3 demonstrates the lowest energy consumption, while it performs the highest AoI, likely attributable to its conservative action selection. Overall, the proposed SAC-TLA successfully establishes an optimal trade-off between energy consumption and AoI, thereby highlighting its robustness in dynamic environments. 

\section{Conclusion}
\par In this paper, we investigated a UAV-assisted AoI-sensitive data forwarding system based on distributed beamforming in IoT. We initially considered a typical scenario in which the sensor data from SNs could not be directly collected by the BS due to long-range transmission distances and intricate terrestrial network conditions. Consequently, UAVs were deployed to collect the data and relay it to the BS by using distributed beamforming  in open-area environments. Subsequently, we formulated a joint optimization problem aimed at minimizing the time-average AoI of  SNs and energy consumption of UAVs. To address this, we proposed a SAC-TLA algorithm, enhanced with several modifications. Simulation results demonstrated that the proposed SAC-TLA algorithm outperformed various benchmark algorithms, especially with regard to reducing AoI and energy consumption.

\section*{Acknowledgement}
\par This study is supported in part by the National Natural Science Foundation of China (62172186, 62272194, 62471200), in part by the Science and Technology Development Plan Project of Jilin Province (20230201087GX), in part by the Postdoctoral Fellowship Program of China Postdoctoral Science Foundation (GZC20240592), in part by the China Postdoctoral Science Foundation General Fund (2024M761123), and in part by the Scientific Research Project of Jilin Provincial Department of Education (JJKH20250117KJ).

\bibliographystyle{IEEEtran}  
\bibliography{aoi}     
\vspace{12pt}

\end{document}